\documentclass[letterpaper, 10 pt, conference]{ieeeconf}  

\IEEEoverridecommandlockouts                              
\usepackage[utf8]{inputenc}
\usepackage[T1]{fontenc}

\usepackage{graphicx}
\usepackage{amsmath}
\usepackage{booktabs}
\usepackage{amssymb}
\usepackage{multirow}
\usepackage{float}
\usepackage{makecell}
\usepackage{booktabs}
\usepackage{array}
\usepackage{tabularx}
\usepackage{placeins}

\title{\LARGE \bf
LLM-Driven 3D Scene Generation of Agricultural Simulation Environments}

\author{Arafa Yoncalik$^{1}$, Wouter Jansen$^{1, 2, 3}$, Nico Huebel$^{1}$, Mohammad Hasan Rahmani$^{1, 2}$ and Jan Steckel$^{1, 2}$
\thanks{$^{1}$All authors are with with Faculty of Applied Engineering – Dept. of Electronics and ICT, University of Antwerp, 2000 Antwerp, Belgium
}%
\thanks{$^{2}$Wouter Jansen, Mohammad Hasan Rahmani and Jan Steckel are with Flanders Make Strategic Research Centre, 3920 Lommel, Belgium}
\thanks{$^{3}$Corresponding author: Wouter Jansen (wouter.jansen@uantwerpen.be)}
}%
\begin{document}

\maketitle
\thispagestyle{empty}
\pagestyle{empty}

\begin{abstract}
Procedural generation techniques in 3D rendering engines have revolutionized the creation of complex environments, reducing reliance on manual design. Recent approaches using Large Language Models (LLMs) for 3D scene generation show promise but often lack domain-specific reasoning, verification mechanisms, and modular design. These limitations lead to reduced control and poor scalability. This paper investigates the use of LLMs to generate agricultural synthetic simulation environments from natural language prompts, specifically to address the limitations of lacking domain-specific reasoning, verification mechanisms, and modular design. A modular multi-LLM pipeline was developed, integrating 3D asset retrieval, domain knowledge injection, and code generation for the Unreal rendering engine using its API. This results in a 3D environment with realistic planting layouts and environmental context, all based on the input prompt and the domain knowledge. To enhance accuracy and scalability, the system employs a hybrid strategy combining LLM optimization techniques such as few-shot prompting, Retrieval-Augmented Generation (RAG), finetuning, and validation. Unlike monolithic models, the modular architecture enables structured data handling, intermediate verification, and flexible expansion. The system was evaluated using structured prompts and semantic accuracy metrics. A user study assessed realism and familiarity against real-world images, while an expert comparison demonstrated significant time savings over manual scene design. The results confirm the effectiveness of multi-LLM pipelines in automating domain-specific 3D scene generation with improved reliability and precision. Future work will explore expanding the asset hierarchy, incorporating real-time generation, and adapting the pipeline to other simulation domains beyond agriculture.
\end{abstract}

\begin{figure*}[!ht]
    \centering
    \includegraphics[width=0.8\textwidth]{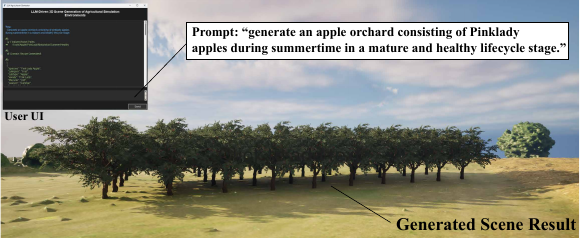}
    \caption{An example prompt-to-scene generation. The interface takes a natural language prompt and outputs a procedurally generated agricultural environment in Unreal Engine.}
    \label{fig:teaser}
\end{figure*}

\section{Introduction}
\label{sec:introduction}

In agriculture, traditional data collection is often constrained by seasonal cycles, weather conditions, and the high costs of physical experimentation. Yet data plays a critical role in advancing precision farming, automation, and informed decision-making \cite{https://doi.org/10.1093/aepp/ppx056}. For instance, data-driven models have been shown to enhance quality assurance and anomaly detection in agricultural monitoring systems \cite{10815930}. Moreover, simulation environments offer a viable alternative to field-based data collection, enabling the generation of synthetic datasets under controlled conditions and varying scenarios, which are essential for training and evaluation of smart systems \cite{shamshiri2018simulation}.

Despite these advances, the creation of agricultural simulation environments remains a time-intensive process. Manually constructing virtual farmlands requires detailed modeling of terrain, crop spacing, growth stages, and environmental variation, which is demanding and difficult to scale. While Procedural Content Generation (PCG) has introduced automation into 3D scene design, existing tools often fall short in addressing domain-specific rules essential to agricultural realism. Crop-specific characteristics, such as species type, lifecycle stage, seasonal growth behavior, and spatial layout rules, require tailored logic that is difficult to generalize using standard PCG methods \cite{emilien2012procedural}. Managing complex datasets that encapsulate agricultural knowledge further complicates this process, requiring the integration of domain-aware automation \cite{Williamson2021}. The introduction of procedural approaches in 3D engines has significantly accelerated the creation of complex virtual environments, reducing reliance on manual design. This trend has been further driven by the widespread availability of accessible technologies such as Unity, Unreal Engine, and Blender, which have broadened access to 3D content creation across domains beyond gaming, including education, architecture, and simulation \cite{Morse03072021}.

Large Language Models (LLMs) present the opportunity to bridge this gap. Their ability to understand natural language and generate structured outputs opens the door to automating scene generation in 3D engines \cite{sun20243dgptprocedural3dmodeling, yang2024llplace3dindoorscene}. However, current single-LLM approaches face several limitations. They often lack access to structured agricultural knowledge such as planting rules, lifecycle stages, seasonal characteristics, or spatial configurations. These systems also struggle to validate their outputs, making them prone to hallucinations or unrealistic scenes. Moreover, monolithic designs offer limited modularity, which restricts their ability to scale across diverse scenarios or adapt to domain-specific constraints. As a result, such systems fall short in generating realistic, consistent, and interpretable 3D agricultural simulations.

In 3D engines like Unreal, environments are constructed from individual 3D objects called assets. This paper investigates a full LLM-based approach for generating agricultural simulation environments, focusing on the proper integration of LLMs across the entire pipeline. Inspired by recent work showing the potential of LLM-driven modular pipelines in other domains \cite{sun20243dgptprocedural3dmodeling, yang2024llplace3dindoorscene}, our system embeds expert knowledge directly into the generation process to bridge the gap between high-level intent and low-level procedural logic. The proposed architecture decomposes the task into three coordinated components: (1) asset retrieval from a structured hierarchy, (2) domain knowledge integration via a hybrid Retrieval-Augmented Generation (RAG) system, and (3) Python code generation that drives environment rendering through the Unreal Engine API. Multiple LLM optimization techniques are combined, including few-shot prompting, finetuning, RAG, and LLM-based output validation, to improve accuracy, flexibility, and generalization. It is important to note that this work does not claim LLMs to be inherently superior to alternative methods such as SQL-based retrieval or purely procedural generation. Instead, it aims to evaluate and refine how LLMs can be effectively integrated into a structured pipeline for this task, identifying where they add value, where they fall short, and how they might complement other approaches. The system is evaluated quantitatively, through semantic accuracy metrics, and qualitatively, through a user study and an expert timing comparison with manual design.

\section{Related Work}
\label{sec:relatedwork}

\subsection{Procedural Generation in 3D Simulations}

Procedural generation refers to the algorithmic creation of data, often in real-time, to produce complex, scalable environments. It has become a cornerstone technology in 3D simulations and offers significant efficiency gains by reducing manual design efforts~\cite{emilien2012procedural, 10.1145/2556288.2557341}. These techniques have been adopted in domains ranging from gaming and architecture to urban planning and scientific visualization~\cite{togelius_et_al:DFU.Vol6.12191.61}. Titles such as \textit{Minecraft} exemplify the use of procedural algorithms to create expansive, diverse worlds with minimal manual intervention~\cite{dahrn-2021}. Modern PCG has evolved from static rule-based systems to dynamic, machine learning-enhanced methods capable of embedding functional properties such as soil variability, irrigation zones, and crop interactions~\cite{10.1145/3664647.3681129, 9954643, bontrager2021learninggeneratelevels}.

\subsection{Generative AI and Multi-LLM Architectures}

Generative AI, particularly LLMs, has revolutionized natural language processing, code generation, and creative workflows. LLMs excel at transforming unstructured prompts into structured outputs, making them ideal for scene generation tasks that require both linguistic understanding and procedural logic~\cite{10298329}. In recent developments such as 3D-GPT, LLMs were successfully applied to procedural scene creation using frameworks such as Blender and Infinigen, demonstrating the viability of prompt-based 3D generation \cite{sun20243dgptprocedural3dmodeling}. Multi-LLM architectures extend this idea by distributing specialized responsibilities across different models \cite{wu2023autogenenablingnextgenllm}. This modular structure supports task-specific optimization and reduces semantic drift, token limitations, and performance bottlenecks~\cite{shen2024smallllmsweaktool, chang2024mainragmultiagentfilteringretrievalaugmented}. Prompt-based LLM pipelines offer a compelling balance between the accuracy of PCG algorithms and the flexibility of natural language interfaces. Unlike manual design, which offers full control but is time-intensive, or rigid PCG scripts, which require significant setup, prompt-based systems allow users to express scene goals intuitively. The system then automates the generation process while ensuring adherence to domain logic and scene consistency.

\subsection{Applications in Agricultural Simulation}

Simulation plays an increasingly important role in modern agriculture. 3D environments are used for training autonomous systems, planning layouts, testing planting strategies, and visualizing environmental impacts~\cite{CHIPANSHI199957, STOCKLE2003289, 10844275}. AI-driven simulations train robots for tasks such as harvesting and weeding, reducing the need for real-world trial-and-error~\cite{chowdhuryai}. Despite these advances, procedural generation in agriculture is still underutilized. Existing tools often lack the capability to incorporate dynamic, rule-based agricultural knowledge such as optimal planting density, disease states, or growth curves across seasons.

\begin{figure*}[t]
  \centering
  \includegraphics[width=\textwidth]{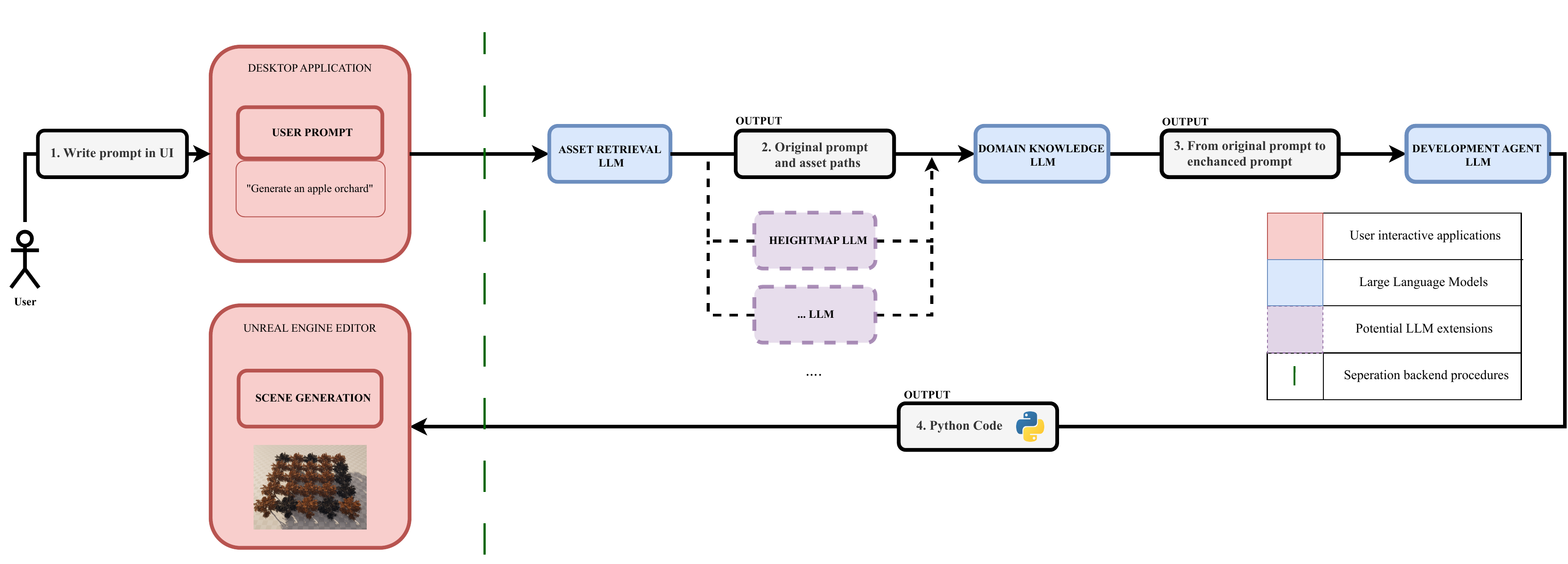}
  \caption{High-level architecture of the multi-LLM system.}
  \label{fig:system-architecture}
\end{figure*}

\section{Methods}
\label{sec:methods}

\subsection{System Overview}

The system developed in this paper is a modular, multi-LLM pipeline designed to convert natural language prompts into procedurally generated 3D agricultural scenes in Unreal Engine (see Figure~\ref{fig:system-architecture}). The architecture divides the overall task into three logically distinct stages: (1) asset path retrieval, (2) domain knowledge injection, and (3) code generation. Each stage is powered by a hybrid LLM-based mechanism, enabling a layered approach that reduces task complexity per model and increases reliability.

The pipeline begins with a user prompt, typically a natural language description such as “Generate an apple orchard of the type Pink Lady during summer in a mature growth stage.” This input is passed to a standalone Desktop Application, where the user communicates through a graphical interface. Once submitted, the backend activates the three-stage LLM pipeline.

First, the Asset Retrieval LLM parses the input and returns structured asset paths from a predefined hierarchy. Next, the Domain Knowledge LLM enriches this with agricultural metadata retrieved using a RAG technique \cite{gao2024retrievalaugmentedgenerationlargelanguage}. Finally, the Development Agent LLM generates Python code based on the enhanced prompt and metadata. This code is then executed inside the Unreal Engine Editor, which the user can also interact with to view the resulting 3D scene (see Figure~\ref{fig:teaser}). The modular architecture allows optimization techniques at each stage, such as few-shot prompting, subquery decomposition, prompt tuning, and validation.

\subsection{Asset Hierarchy Design}

To ensure consistent asset referencing, a structured directory hierarchy was implemented for all agricultural 3D models used in the system. The hierarchy separates content into two main categories \textit{Fruits} and \textit{Vegetables} each of which contains multiple crop types. Each crop type is further subdivided by specific \textit{varieties}, \textit{growth stages}, \textit{seasonal appearance}, and \textit{health conditions}. This hierarchical structure enables the system to map natural language prompts to precise asset paths, ensuring visual consistency and domain fidelity in generated environments.

The hierarchy was generated programmatically and, for the purposes of this study, was limited to a predefined test dataset containing a representative subset of crops and hierarchy elements. This selection was made to balance domain coverage with feasibility for evaluation, while still enabling diverse scene generation. It spans fruits (e.g., apple, banana, cherry), vegetables (e.g., carrot, lettuce, tomato), multiple growth stages (vegetative, reproductive, maturation), seasonal appearances (fall, winter, summer, spring), and health states (healthy, ill). In total, the structure allows for 672 asset path combinations, supporting a wide variety of simulation scenarios. All assets are stored in a format compatible with Unreal Engine, and their paths are embedded in a semantic retrieval system to enable intelligent selection via LLMs. The asset hierarchy also plays a critical role in the system’s retrieval and code generation phases, acting as the backbone for ensuring that procedural scenes are populated with appropriate visual representations based on the user’s prompt.

\subsection{Asset Retrieval LLM}

To convert natural language prompts into valid asset paths, several LLM optimization strategies were evaluated, including few-shot prompting, finetuning, and RAG. Based on performance across varying prompt types, a hybrid method was selected. This approach combines subquery decomposition, semantic search, and GPT-based refinement to balance flexibility with strict adherence to the asset hierarchy.

The process begins by parsing user prompts into subqueries using GPT-4 \cite{openai2024gpt4technicalreport}, especially in multi-field scenarios. These subqueries are then standardized according to the system’s asset hierarchy. For example, loosely defined terms like “early stage” or “flowering” are normalized to lifecycle categories such as \texttt{Vegetative}, \texttt{Reproductive}, or \texttt{Maturation}. 

Normalized subqueries are embedded using OpenAI’s \texttt{text-embedding-3-small} model and compared against a FAISS index of known asset paths. This semantic search retrieves the closest matches based on meaning, not exact wording, enabling flexible interaction while ensuring structural consistency \cite{10392009}. A final GPT-4 validation step ensures that retrieved paths match the user’s intent in terms of variety, lifecycle, season, and health state. It also enforces internal consistency, such as matching seasonal context across multiple fields.

\subsection{Domain Knowledge Integration via RAG}

Following asset retrieval, the system enriches each asset path with structured agricultural metadata via a custom RAG pipeline. This ensures that the final scene generation reflects realistic properties such as crop spacing, growth behavior, and seasonal effects.

A FAISS index stores embeddings of domain knowledge entries, each a JSON object describing a specific crop configuration, including variety, lifecycle, season, health state, and properties like height, density, disease susceptibility, irrigation, and rendering effects. Metadata fields parsed from the asset path (e.g. category, variety, season) are combined into a descriptor string (e.g. “healthy young Pink Lady apple in fall”), embedded using the same model as the index, and matched via top-k semantic search. A secondary filter ensures retrieved entries fully align with the asset path, including fuzzy variety matching. If no match is found, the system logs the failure for fallback handling.

The final output is a structured recipe of enriched JSON entries containing both agricultural metadata and scene-specific attributes (e.g., model tags, scaling, environmental settings). This ensures each generated scene is not only visually coherent but also agronomically accurate and context-aware.

\subsection{Code Generation LLM and Unreal Engine Integration}

The final stage of the pipeline transforms the domain-enriched scene recipe into executable Python code that generates a 3D environment in Unreal Engine. This is accomplished through a specialized Code Generation LLM, which has been optimized via prompt engineering, structural constraints, and custom instruction finetuning to produce Python scripts that are both syntactically correct and compatible within Unreal Engine.

The LLM takes three inputs: the original user prompt, validated asset paths, and the enriched scene recipe retrieved via RAG. It then produces a structured Python script that instantiates and places 3D assets based on domain metadata, such as lifecycle stages, spacing guidelines, and seasonal context. The code includes parameters like model references, row and tree spacing, scaling, rotation, and environmental behaviors. Scripts follow a modular layout, with clear functions for scene setup, looping through domain objects, and applying placement and attachment logic.

To ensure correctness, a final validation step checks for:
\begin{itemize}
    \item Missing or malformed asset paths,
    \item Unreal-specific library errors (e.g., attachment rules or constructor calls),
    \item Mismatches between domain metadata and scene parameters (e.g., incorrect scale).
\end{itemize}

The LLM was finetuned on a curated dataset of prompt–script pairs collected during development, which significantly improved output structure and reduced hallucinations. By combining flexible prompt-to-code generation with deterministic rendering in Unreal Engine, this stage closes the loop between natural language input and visual output.

\section{Results and Evaluation}
\label{sec:resultsandevaluation}

\noindent 
The evaluation covers the complete pipeline, starting with asset retrieval, followed by domain knowledge alignment, and concluding with Unreal Python code generation. Beyond these core stages, we also compare the modular multi-LLM architecture against a single-LLM baseline and present two human-centered studies: one on user-perceived scene quality and another on expert time efficiency. Quantitative metrics include accuracy, precision, recall, F1 score, and Top-$k$ retrieval accuracy, while qualitative assessments capture visual realism, semantic correctness, and scalability considerations.

\begin{table}[t]
\centering
\caption{Evaluation of asset retrieval across prompt types}
\label{tab:retrieval-comparison}
\resizebox{\linewidth}{!}{%
\begin{tabular}{llcccc}
\toprule
\textbf{Prompt Type} & \textbf{Metric} & \textbf{Few-Shot} & \textbf{Finetuning} & \textbf{RAG} & \textbf{Hybrid} \\
\midrule
Single-field (detailed) & Accuracy & 100\% & 100\% & 97\% & 98\% \\
\addlinespace
Single-field (generic)  & Accuracy & 83\%  & 66\%  & 87\% & 71\% \\
\addlinespace
\multirow[c]{3}{*}{Multi-field (generic)} 
    & Precision & 74\% & 63\% & 74\% & N/A \\
    & Recall    & 6\%  & 89\% & 8\%  & N/A \\
    & F1 Score  & 10\% & 74\% & 15\% & N/A \\
\bottomrule
\end{tabular}%
}
\end{table}

\subsection{Asset Retrieval Evaluation}

While LLMs are commonly assessed using standardized benchmarks such as MMLU, HellaSwag, or HumanEval~\cite{ivanov2024aibenchmarksdatasetsllm}, these benchmarks are designed for general purpose reasoning or programming tasks. They do not capture the domain-specific requirements involved in asset retrieval, agricultural metadata alignment, or 3D scene generation. To address this, a custom evaluation framework was developed to assess performance across the three key stages of the pipeline: asset path retrieval, domain knowledge integration, and code generation. This framework uses manually constructed ground truth sets, accuracy-based metrics, and visual validation within Unreal Engine to assess real-world applicability. To evaluate asset path retrieval performance, a benchmark of 100 natural language prompts was created and categorized as follows:

\begin{itemize}
    \item \textbf{Single-field, detailed}: e.g., “Generate a healthy Pink Lady apple orchard in summer.”
    \item \textbf{Single-field, generic}: e.g., “Generate an apple field.”
    \item \textbf{Multi-field, generic}: e.g., “Generate some fruit and vegetable fields.”
\end{itemize}

For each prompt, a ground truth set of expected asset paths was manually constructed. Performance was evaluated across four configurations: few-shot prompting, finetuning, RAG, and a hybrid approach. Each was evaluated using 100 prompts across different categories (detailed vs. generic, single-field vs. multi-field). The metrics assessed include accuracy, precision, recall, and F1 score.  For the hybrid system, only accuracy was measured, as it is explicitly designed to return only the minimal necessary asset paths per prompt, rather than an exhaustive list of all valid options. As a result, metrics like precision and recall, which assume comprehensive retrieval (i.e., returning all correct options), become less informative for this evaluation. These results can be found in Table \ref{tab:retrieval-comparison}.

These results highlight the strengths and trade-offs of each approach. Few-shot prompting and finetuning achieved perfect accuracy for detailed single-field prompts. Finetuning demonstrated strong recall and F1 in multi-field queries, showing its ability to retrieve broader asset sets. RAG achieved high precision but lower recall, indicating that it retrieved fewer but often correct paths. The hybrid approach focused on high-precision retrieval for low-ambiguity queries by minimizing false positives, though precision and recall metrics were not computed due to its constrained output design.

\begin{table}[htbp]
\centering
\caption{Domain Knowledge Retrieval Accuracy (Exact Match)}
\label{tab:domain-knowledge}
\small
\setlength{\tabcolsep}{10pt}
\renewcommand{\arraystretch}{1}
\begin{tabular}{lccc}
\toprule
\makecell{\textbf{Method}} &
\makecell{\textbf{Top-1}\\\textbf{Accuracy}} &
\makecell{\textbf{Top-2}\\\textbf{Recall}} &
\makecell{\textbf{Top-3}\\\textbf{Recall}} \\
\midrule
RAG           & 71\%  & 98\% & 100\% \\
Hybrid Filter & 82\%  & 96\% & 100\% \\
\bottomrule
\end{tabular}
\end{table}

\subsection{Domain Knowledge Evaluation}

To evaluate the accuracy and consistency of domain knowledge retrieval, two methods were tested: a baseline RAG approach and an improved hybrid approach combining RAG with strict metadata filtering. Both systems retrieved relevant JSON metadata entries from a FAISS index based on the asset paths selected during the previous stage. The full results can be found in Table \ref{tab:domain-knowledge}.

The first method, \textit{RAG}, directly embedded the asset path as text and performed semantic similarity search. While simple, this method lacked deterministic control over field-specific attributes such as season or lifecycle stage. A Top-1 search achieved only 71\% accuracy in exact metadata matches. This often led to scene generation errors, such as assigning summer properties to a spring apple orchard. 

In contrast, the \textit{hybrid approach} first parsed asset paths into structured metadata (e.g., category, subtype, variety, lifecycle, season, health), embedded this metadata, and then retrieved the top-k semantically close candidates. A strict post-filtering stage then validated whether each candidate’s metadata matched the query exactly. This significantly outperformed RAG in Top-1 accuracy while maintaining the same Top-3 coverage. Even at Top-3 retrieval, the hybrid system queried less than 0.5\% of the total database, demonstrating high efficiency.

The hybrid method improved consistency across all prompt types by avoiding mismatched fields (e.g., incorrect seasons) and enabling more precise scene recipes. This accuracy is critical because errors in this stage propagate directly into code generation, affecting the visual and semantic fidelity of the final simulation environment.

\begin{table}
\centering
\caption{Development Agent Code Generation Results (10 prompts)}
\label{tab:code-generation-results}
\small
\setlength{\tabcolsep}{6pt} 
\renewcommand{\arraystretch}{1}
\begin{tabular}{lcccc}
\toprule
\textbf{Prompt Type} &
\makecell{\textbf{Execu-}\\\textbf{tability}} &
\makecell{\textbf{Correct}\\\textbf{Paths}} &
\makecell{\textbf{Domain}\\\textbf{Match}} &
\makecell{\textbf{Visual}\\\textbf{Accuracy}} \\
\midrule
Single-field & 100\% & 100\% & 100\% & 100\% \\
Multi-field  & 100\% & 100\% & 100\% & 70\% \\
\bottomrule
\end{tabular}
\end{table}

\subsection{Code Generation Evaluation}

This section evaluates the performance of the multi-LLM pipeline for generating Unreal Engine Python scripts based on structured scene recipes. 

\noindent All results reported here are produced by the modular system described earlier, with separate LLMs for asset retrieval, domain knowledge enrichment, and code generation. To assess the performance of the development agent, two sets of evaluations were conducted: one using 10 single-field prompts and another using 10 more complex multi-field prompts. Each generated script was assessed on four criteria: whether the generated code was executable within Unreal Engine, the accuracy of asset path utilization, the scene's adherence to domain-specific knowledge such as object spacing, scaling, and structural layout, and, if the scene executed successfully, whether the visual output aligned with the original prompt's intent. The full results can be found in Table \ref{tab:code-generation-results}.

The 10 single-field prompts produced scripts that were executable, referenced correct asset paths, respected domain knowledge properties (e.g., spacing and scaling), and visually matched the user description. Minor visual inaccuracies due to shared asset meshes across seasons were observed, but these did not affect correctness at the code level.

In comparison, the evaluation of 10 multi-field prompts showed similar robustness in execution and metadata handling. All scripts were executable and semantically consistent. However, three outputs exhibited visual limitations: two failed to generate correct spatial relationships between fields (e.g. missing layout offsets), and one spawned a single asset per type rather than full fields.

Figure~\ref{fig:teaser} shows an example of a system-generated agricultural field produced by the pipeline, visualized in Unreal Engine based on a natural language prompt. This illustrates the spatial layout, variety selection, and planting logic applied by the code generation stage.

\subsection{Single vs Multi-LLM Architecture Evaluation}

The results reported in the previous \textit{Code Generation Evaluation} subsection reflect the performance of the proposed modular multi-LLM pipeline, in which code generation is informed by separate stages for asset retrieval and domain knowledge integration. While AI-driven systems usually rely on a single LLM to manage the entire pipeline from natural language interpretation to execution, this paper adopts a modular architecture. The pipeline divides the process into three specialized components: asset path retrieval, domain knowledge enrichment, and Unreal Python code generation. This separation enhances understanding, simplifies debugging, and allows for targeted optimization at each stage.
\newpage
To evaluate the effectiveness of this design choice, a controlled internal comparison was conducted between the modular multi-LLM pipeline and a baseline single-LLM configuration. In the baseline setup, GPT-4 received a raw user prompt along with an embedded task description and few-shot examples. It was expected to infer asset paths, apply domain-specific logic, and generate Unreal-compatible code within a single step. The outputs from both approaches were evaluated using a combination of qualitative and quantitative criteria:

\begin{itemize}
    \item \textbf{Modularity:} The single-LLM setup lacked clear stage separation, making debugging and targeted improvements difficult. The modular pipeline enabled finer control over each step.
    \item \textbf{Scalability:} The single-LLM approach hit prompt length limits with large asset taxonomies, while the multi-LLM system stored the hierarchy in a searchable embedded database.
    \item \textbf{Correctness:} The single-LLM method often hallucinated asset paths or omitted domain metadata. Intermediate validation in the modular pipeline greatly reduced such errors.
    \item \textbf{Flexibility:} The multi-LLM design supported specialized methods like RAG, lifecycle normalization, and code validation loops that are harder to combine in a single prompt.
\end{itemize}

To ground this comparison, we applied the same evaluation protocol used for the code generation stage: ten single-field and ten multi-field prompts were used to assess scene generation quality. Each result was scored on code executability, asset path correctness, domain knowledge application, and visual accuracy. These results can be found in Table~\ref{tab:single-llm-results}.

\begin{table}[t]
\centering
\caption{Single-LLM results (10 prompts)}
\label{tab:single-llm-results}
\small
\setlength{\tabcolsep}{10pt}
\renewcommand{\arraystretch}{1}
\begin{tabular}{lccc}
\toprule
\makecell{\textbf{Prompt} \textbf{Type}} &
\makecell{\textbf{Execu-}\\\textbf{tability}} &
\makecell{\textbf{Domain}\\\textbf{Match}} &
\makecell{\textbf{Visual}\\\textbf{Accuracy}} \\
\midrule
Single-field & 70\%  & 80\% & 100\% \\
Multi-field  & 100\% & 70\% & 90\%  \\
\bottomrule
\end{tabular}
\end{table}

An observed issue with the single-LLM system was its failure to automatically format asset paths correctly, particularly omitting necessary Unreal Engine path prefixes such as \texttt{/Game/} and file suffixes like \texttt{.fbx}. Without manual intervention, none of the scripts executed. Once corrected, the system could produce working outputs, but domain metadata such as crop density and field spacing often remained inaccurate. By contrast, the multi-LLM system with dedicated stages for asset retrieval and metadata alignment generated executable and visually accurate results without manual path manipulation. The modular pipeline also exhibited more flexible and diverse code structures, as opposed to the repetitive patterns observed in the single-LLM outputs. These findings support the architectural choice to separate the generation process into stages and shows better results for real-world deployment in large asset ecosystems.

\subsection{Interpretation and Scalability Considerations}

While standard metrics like accuracy, precision, and F1 score offer valuable insight into prompt handling and asset matching, they do not fully capture the system’s real-world utility. For example, the high accuracy of few-shot prompting in single-field prompts is partly due to controlled testing conditions: the asset hierarchy contained only 672 combinations, small enough to fit entirely in the system prompt, giving GPT-4 complete visibility over all valid paths.

This setup is not scalable. As asset paths grow into the thousands, prompt size and context limits make few-shot prompting insufficient. Finetuning improved format consistency but lacked the precision required to return exact asset paths, essential for Unreal Engine scene generation. This motivated a shift to RAG, which enabled accurate and scalable semantic retrieval from an embedded index of asset paths without modifying the base model. However, RAG alone lacked the reasoning to map vague or multi-field prompts reliably. The hybrid approach addressed this by combining sub-query decomposition, normalization, and semantic validation to better align user intent with database retrieval.

It is also important to highlight that precision, recall, and F1 scores are only practical in our case because the dataset is relatively small and the full set of valid answers is known. For significantly larger asset hierarchies, such evaluations may become infeasible due to output token limits and ambiguity in defining exhaustive ground truths.

Finally, model performance evolves over time; future results may differ even with identical tests. Our design therefore prioritizes adaptability, modularity, and correctness over short-term benchmark gains.

\subsection{User Evaluation and Results}

A user study was conducted with 10 participants to evaluate the 
human-perceived quality of the generated scenes. Each participant completed 
three evaluations, resulting in 30 scene assessments. The goal was to 
determine whether the multi-LLM pipeline could accurately interpret natural 
language descriptions and produce visually realistic agricultural environments. 
In each evaluation, participants were shown a real-world image of an 
agricultural field and asked to describe it freely. This description was then 
used as input to the system, which generated a corresponding 3D scene in 
Unreal Engine. A screenshot of the generated scene was presented alongside the 
original image for evaluation. Participants rated the system output using the 
following questions:

\begin{itemize}
    \item \textbf{Q1.} To what extent does this generated scene match what you described? (1–10)
    \item \textbf{Q2.} How similar is this scene to the original image (excluding terrain)? (1–10)
    \item \textbf{Q3.} How realistic is the generated environment? (1 = Not realistic, 5 = Extremely realistic)
    \item \textbf{Q4.} Are species, spacing, and layout agriculturally accurate? (Self-rated expertise: 1–10)
\end{itemize}

Participants also responded to two open-ended questions on missing elements 
and general suggestions. 

\subsubsection{Results and Interpretation}

Average scores from the 30 evaluations are shown in Table~\ref{tab:user-eval}. 
The "Agri Expertise" score reflects participants' self-assessed expertise rather 
than the objective quality of the layout. Participants generally perceived the 
system’s scene generation as moderately accurate in capturing their descriptions. 
However, realism and visual similarity scores were lowered due to repetitive 
assets and the absence of terrain or soil elements. As confirmed in open-ended 
feedback, many visual mismatches were due to generic 3D models or missing 
environmental variation. Future inclusion of terrain and ground generation is 
expected to significantly improve visual coherence and user satisfaction.

\begin{table}[H]
\centering
\caption{User Evaluation Summary (n = 30)}
\label{tab:user-eval}
\begin{tabular}{lc}
\toprule
\textbf{Metric} & \textbf{Average Score (1–10)} \\
\midrule
Prompt Match (Q1)      & 6.77 \\
Visual Similarity (Q2) & 5.50 \\
Scene Realism (Q3)     & 5.54 \\
Agri Expertise (Q4)    & 3.70 \\
\bottomrule
\end{tabular}
\end{table}

\subsection{Expert Evaluation}

\begin{figure}[htbp]
    \centering
    \includegraphics[width=1\linewidth]{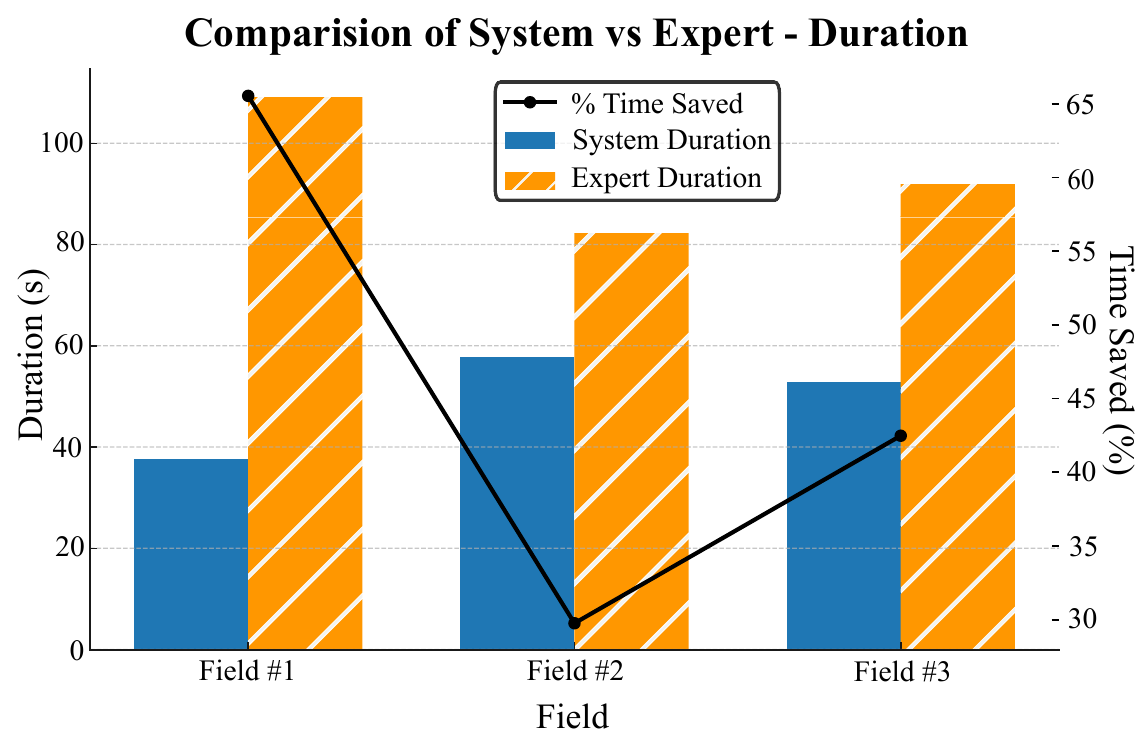}
    \caption{Comparison of generation time: System vs Expert (Single-Field Scenes).}
    \label{fig:expert-time-bar}
\end{figure}

To benchmark the system’s performance against traditional workflows, an expert evaluation was conducted involving three professionals experienced in Unreal Engine and 3D scene construction. The goal was to compare manual and automated scene generation in terms of time efficiency. Each expert was asked to manually recreate three single-field agricultural scenes based on natural language prompts previously used by the system. For fairness, the correct asset paths were provided in advance, so the evaluation focused strictly on placement, scaling, and spacing, excluding time spent searching through the asset hierarchy. The same scenes were first generated using the multi-LLM pipeline, and the total generation time was recorded. The results are shown in Table~\ref{tab:expert-time} and visualized in Figure~\ref{fig:expert-time-bar}.

\begin{table}[htbp]
\centering
\caption{System vs Expert Scene Generation Time (Single-Field)}
\label{tab:expert-time}
\resizebox{\linewidth}{!}{%
\begin{tabular}{lcccc}
\toprule
\textbf{Field} & \textbf{System (s)} & \textbf{Expert (s)} & \textbf{Saved (s)} & \textbf{\% Faster} \\
\midrule
Scene 1 & 37.61 & 109.13 & 71.52 & 65.54\% \\
Scene 2 & 57.75 & 82.22  & 24.47 & 29.76\% \\
Scene 3 & 52.83 & 91.85  & 39.02 & 42.47\% \\
\bottomrule
\end{tabular}
}
\end{table}

On average, the system required only 49 seconds per scene, while experts took 94.4 seconds, almost twice as long. This shows a substantial efficiency gain. Given the simplicity of single-field scenes, it is reasonable to expect that the time savings would be even greater in more complex, multi-field scenarios. As the system evolves to include features such as terrain generation, seasonal weather, and foliage variation, this time advantage is likely to expand even further.

\section{Discussion \& Conclusion}
\label{sec:discussion}

This paper explored a structured method for generating 3D agricultural environments in Unreal Engine using a modular system powered by LLMs. These results highlight the value of combining generative AI with structured domain logic, providing a foundation for developing interactive, scalable simulation tools across domains beyond agriculture.
By dividing the process into asset retrieval, domain knowledge enrichment, and code generation, the system addresses key challenges in aligning natural language with precise scene composition. 

While the proposed multi-LLM pipeline shows promise in generating agriculturally accurate 3D environments, several limitations remain. The system relies on static, pre-defined assets, with realism constrained by asset quality and metadata. It lacks procedural generation and dynamic features such as crop growth or environmental interactions.

\noindent Despite hybrid asset retrieval, ambiguous or under-specified prompts can still cause inconsistencies or hallucinated metadata, reflecting the need for broader and more balanced knowledge coverage. Reliance on external API services also introduces latency, costs, and internet dependency. Evaluation was further limited by a small participant sample and narrow scene variety. 

From a code generation perspective, the system consistently applies placement, scaling, and row-spacing rules, but does not yet incorporate more complex agricultural attributes such as seasonal lighting effects, foliage density, or disease visualizations. These omissions are partly due to limitations in asset modularity, highlighting the need for richer and more flexible asset libraries.

Future work should expand domain knowledge to include weather, soil, and biological variation, integrate procedural terrain and environment generation, and enable interactive editing or selective regeneration. Beyond agriculture, the modular multi-LLM architecture could also extend to domains such as urban planning, forestry, and environmental conservation.

\bibliographystyle{ieeetr}
\bibliography{main}

@article{https://doi.org/10.1093/aepp/ppx056,
author = {Coble, Keith H and Mishra, Ashok K and Ferrell, Shannon and Griffin, Terry},
title = {Big Data in Agriculture: A Challenge for the Future},
journal = {Applied Economic Perspectives and Policy},
volume = {40},
number = {1},
pages = {79-96},
doi = {https://doi.org/10.1093/aepp/ppx056},
url = {https://onlinelibrary.wiley.com/doi/abs/10.1093/aepp/ppx056},
eprint = {https://onlinelibrary.wiley.com/doi/pdf/10.1093/aepp/ppx056},
year = {2018}
}

@misc{bontrager2021learninggeneratelevels,
      title={Learning to Generate Levels From Nothing}, 
      author={Philip Bontrager and Julian Togelius},
      year={2021},
      eprint={2002.05259},
      archivePrefix={arXiv},
      primaryClass={cs.AI},
      url={https://arxiv.org/abs/2002.05259}, 
}

@misc{wu2023autogenenablingnextgenllm,
      title={AutoGen: Enabling Next-Gen LLM Applications via Multi-Agent Conversation}, 
      author={Qingyun Wu and Gagan Bansal and Jieyu Zhang and Yiran Wu and Beibin Li and Erkang Zhu and Li Jiang and Xiaoyun Zhang and Shaokun Zhang and Jiale Liu and Ahmed Hassan Awadallah and Ryen W White and Doug Burger and Chi Wang},
      year={2023},
      eprint={2308.08155},
      archivePrefix={arXiv},
      primaryClass={cs.AI},
      url={https://arxiv.org/abs/2308.08155}, 
}

@misc{chang2024mainragmultiagentfilteringretrievalaugmented,
      title={MAIN-RAG: Multi-Agent Filtering Retrieval-Augmented Generation}, 
      author={Chia-Yuan Chang and Zhimeng Jiang and Vineeth Rakesh and Menghai Pan and Chin-Chia Michael Yeh and Guanchu Wang and Mingzhi Hu and Zhichao Xu and Yan Zheng and Mahashweta Das and Na Zou},
      year={2024},
      eprint={2501.00332},
      archivePrefix={arXiv},
      primaryClass={cs.CL},
      url={https://arxiv.org/abs/2501.00332}, 
}

@ARTICLE{10844275,
  author={Noda, Shintaro and Kogoshi, Masayuki and Iijima, Wataru},
  journal={IEEE Access}, 
  title={Robot Simulation on Agri-Field Point Cloud With Centimeter Resolution}, 
  year={2025},
  volume={13},
  number={},
  pages={14404-14416},
  doi={10.1109/ACCESS.2025.3530967}}

@misc{openai2024gpt4technicalreport,
      title={GPT-4 Technical Report}, 
      author={OpenAI and others},
      year={2024},
      eprint={2303.08774},
      archivePrefix={arXiv},
      primaryClass={cs.CL},
      url={https://arxiv.org/abs/2303.08774}, 
}

@ARTICLE{10815930,
  author={Gkountakos, Konstantinos and Ioannidis, Konstantinos and Demestichas, Konstantinos and Vrochidis, Stefanos and Kompatsiaris, Ioannis},
  journal={IEEE Access}, 
  title={A Comprehensive Review of Deep Learning-Based Anomaly Detection Methods for Precision Agriculture}, 
  year={2024},
  volume={12},
  number={},
  pages={197715-197733},
  doi={10.1109/ACCESS.2024.3522248}}

@article{shamshiri2018simulation,
  author    = {Shamshiri, Redmond Ramin and Hameed, Ibrahim A. and Pitonakova, Lenka and Weltzien, Cornelia and Balasundram, Siva K. and Yule, Ian J. and Grift, Tony E. and Chowdhary, Girish},
  title     = {Simulation software and virtual environments for acceleration of agricultural robotics: Features highlights and performance comparison},
  journal   = {International Journal of Agricultural and Biological Engineering},
  volume    = {11},
  number    = {4},
  pages     = {12--20},
  year      = {2018},
  doi       = {10.25165/j.ijabe.20181104.4032},
  url       = {https://doi.org/10.25165/j.ijabe.20181104.4032}
}

@article{emilien2012procedural,
  author    = {Arnaud Emilien and Adrien Bernhardt and Adrien Peytavie and Marie-Paule Cani and Eric Galin},
  title     = {Procedural generation of villages on arbitrary terrains},
  journal   = {The Visual Computer},
  volume    = {28},
  number    = {6},
  pages     = {809--818},
  year      = {2012},
  month     = {June},
  doi       = {10.1007/s00371-012-0699-7},
  url       = {https://doi.org/10.1007/s00371-012-0699-7}
}

@article{Williamson2021,
  author       = {Williamson, HF and Brettschneider, J and Caccamo, M and Davey, RP and Goble, C and Kersey, PJ and May, S and Morris, RJ and Ostler, R and Pridmore, T and Rawlings, C and Studholme, D and Tsaftaris, SA and Leonelli, S},
  title        = {Data management challenges for artificial intelligence in plant and agricultural research},
  journal      = {F1000Research},
  year         = {2021},
  volume       = {10},
  pages        = {324},
  doi          = {10.12688/f1000research.52204.2},
  pmid         = {36873457},
  pmcid        = {PMC9975417},
  month        = {Apr},
  day          = {27}
}

@article{Morse03072021,
author = {Christopher Morse and},
title = {Gaming Engines: Unity, Unreal, and Interactive 3D Spaces},
journal = {Technology|Architecture + Design},
volume = {5},
number = {2},
pages = {246--249},
year = {2021},
publisher = {Routledge},
doi = {10.1080/24751448.2021.1967068},
URL = { 
        https://doi.org/10.1080/24751448.2021.1967068
},
eprint = { 
        https://doi.org/10.1080/24751448.2021.1967068
}
}

@misc{sun20243dgptprocedural3dmodeling,
      title={3D-GPT: Procedural 3D Modeling with Large Language Models}, 
      author={Chunyi Sun and Junlin Han and Weijian Deng and Xinlong Wang and Zishan Qin and Stephen Gould},
      year={2024},
      eprint={2310.12945},
      archivePrefix={arXiv},
      primaryClass={cs.CV},
      url={https://arxiv.org/abs/2310.12945}, 
}

@misc{yang2024llplace3dindoorscene,
      title={LLplace: The 3D Indoor Scene Layout Generation and Editing via Large Language Model}, 
      author={Yixuan Yang and Junru Lu and Zixiang Zhao and Zhen Luo and James J. Q. Yu and Victor Sanchez and Feng Zheng},
      year={2024},
      eprint={2406.03866},
      archivePrefix={arXiv},
      primaryClass={cs.CV},
      url={https://arxiv.org/abs/2406.03866}, 
}

@inproceedings{10.1145/2556288.2557341,
author = {Smith, Gillian},
title = {Understanding procedural content generation: a design-centric analysis of the role of PCG in games},
year = {2014},
isbn = {9781450324731},
publisher = {Association for Computing Machinery},
address = {New York, NY, USA},
url = {https://doi.org/10.1145/2556288.2557341},
doi = {10.1145/2556288.2557341},
booktitle = {Proceedings of the SIGCHI Conference on Human Factors in Computing Systems},
pages = {917–926},
numpages = {10},
location = {Toronto, Ontario, Canada},
series = {CHI '14}
}

@InCollection{togelius_et_al:DFU.Vol6.12191.61,
  author =	{Togelius, Julian and Champandard, Alex J. and Lanzi, Pier Luca and Mateas, Michael and Paiva, Ana and Preuss, Mike and Stanley, Kenneth O.},
  title =	{{Procedural Content Generation: Goals, Challenges and Actionable Steps}},
  booktitle =	{Artificial and Computational Intelligence in Games},
  pages =	{61--75},
  series =	{Dagstuhl Follow-Ups},
  ISBN =	{978-3-939897-62-0},
  ISSN =	{1868-8977},
  year =	{2013},
  volume =	{6},
  editor =	{Lucas, Simon M. and Mateas, Michael and Preuss, Mike and Spronck, Pieter and Togelius, Julian},
  publisher =	{Schloss Dagstuhl -- Leibniz-Zentrum f{\"u}r Informatik},
  address =	{Dagstuhl, Germany},
  URL =		{https://drops.dagstuhl.de/entities/document/10.4230/DFU.Vol6.12191.61},
  URN =		{urn:nbn:de:0030-drops-43367},
  doi =		{10.4230/DFU.Vol6.12191.61}
}

@phdthesis{dahrn-2021,
  author = {Dahrn, M.},
  title = {The Usage of PCG Techniques Within Different Game Genres},
  year = {2021},
  type = {Dissertation},
  institution = {Malmö University},
  url = {https://urn.kb.se/resolve?urn=urn:nbn:se:mau:diva-46422}
}

@inproceedings{10.1145/3664647.3681129,
author = {Liu, Jia-Hong and Zhang, Shao-Kui and Zhang, Chuyue and Zhang, Song-Hai},
title = {Controllable Procedural Generation of Landscapes},
year = {2024},
isbn = {9798400706868},
publisher = {Association for Computing Machinery},
address = {New York, NY, USA},
url = {https://doi.org/10.1145/3664647.3681129},
doi = {10.1145/3664647.3681129},
booktitle = {Proceedings of the 32nd ACM International Conference on Multimedia},
pages = {6394–6403},
numpages = {10},
location = {Melbourne VIC, Australia},
series = {MM '24}
}

@INPROCEEDINGS{9954643,
  author={Vuletić, Jelena and Polić, Marsela and Orsag, Matko},
  booktitle={2022 International Conference on Smart Systems and Technologies (SST)}, 
  title={Procedural Generation of Synthetic Dataset for Robotic Applications in Sweet Pepper Cultivation}, 
  year={2022},
  volume={},
  number={},
  pages={309-314},
  doi={10.1109/SST55530.2022.9954643}}

@INPROCEEDINGS{10298329,
  author={Gao, Shuzheng and Wen, Xin-Cheng and Gao, Cuiyun and Wang, Wenxuan and Zhang, Hongyu and Lyu, Michael R.},
  booktitle={2023 38th IEEE/ACM International Conference on Automated Software Engineering (ASE)}, 
  title={What Makes Good In-Context Demonstrations for Code Intelligence Tasks with LLMs?}, 
  year={2023},
  volume={},
  number={},
  pages={761-773},
  doi={10.1109/ASE56229.2023.00109}}

@misc{shen2024smallllmsweaktool,
      title={Small LLMs Are Weak Tool Learners: A Multi-LLM Agent}, 
      author={Weizhou Shen and Chenliang Li and Hongzhan Chen and Ming Yan and Xiaojun Quan and Hehong Chen and Ji Zhang and Fei Huang},
      year={2024},
      eprint={2401.07324},
      archivePrefix={arXiv},
      primaryClass={cs.AI},
      url={https://arxiv.org/abs/2401.07324}, 
}

@article{CHIPANSHI199957,
title = {Large-scale simulation of wheat yields in a semi-arid environment using a crop-growth model},
journal = {Agricultural Systems},
volume = {59},
number = {1},
pages = {57-66},
year = {1999},
issn = {0308-521X},
doi = {https://doi.org/10.1016/S0308-521X(98)00082-1},
url = {https://www.sciencedirect.com/science/article/pii/S0308521X98000821},
author = {A.C Chipanshi and E.A Ripley and R.G Lawford}
}

@article{STOCKLE2003289,
title = {CropSyst, a cropping systems simulation model},
journal = {European Journal of Agronomy},
volume = {18},
number = {3},
pages = {289-307},
year = {2003},
note = {Modelling Cropping Systems: Science, Software and Applications},
issn = {1161-0301},
doi = {https://doi.org/10.1016/S1161-0301(02)00109-0},
url = {https://www.sciencedirect.com/science/article/pii/S1161030102001090},
author = {Claudio O. Stöckle and Marcello Donatelli and Roger Nelson}
}

@article{chowdhuryai,
  title={AI-Driven Agricultural Robotics: Advancements and Applications},
  author={Chowdhury, Manojit and Anand, Rohit}
}

@misc{gao2024retrievalaugmentedgenerationlargelanguage,
      title={Retrieval-Augmented Generation for Large Language Models: A Survey}, 
      author={Yunfan Gao and Yun Xiong and Xinyu Gao and Kangxiang Jia and Jinliu Pan and Yuxi Bi and Yi Dai and Jiawei Sun and Meng Wang and Haofen Wang},
      year={2024},
      eprint={2312.10997},
      archivePrefix={arXiv},
      primaryClass={cs.CL},
      url={https://arxiv.org/abs/2312.10997}, 
}

@INPROCEEDINGS{10392009,
  author={Ghadekar, Premanand P. and Mohite, Sahil and More, Omkar and Patil, Praiwal and Sayantika and Mangrule, Shubham},
  booktitle={2023 7th International Conference On Computing, Communication, Control And Automation (ICCUBEA)}, 
  title={Sentence Meaning Similarity Detector Using FAISS}, 
  year={2023},
  volume={},
  number={},
  pages={1-6},
  doi={10.1109/ICCUBEA58933.2023.10392009}}

@misc{ivanov2024aibenchmarksdatasetsllm,
      title={AI Benchmarks and Datasets for LLM Evaluation}, 
      author={Todor Ivanov and Valeri Penchev},
      year={2024},
      eprint={2412.01020},
      archivePrefix={arXiv},
      primaryClass={cs.DC},
      url={https://arxiv.org/abs/2412.01020}, 
}

\end{document}